\documentclass[letterpaper,times]{IONconf}
\usepackage{url}

\usepackage{graphics}
\usepackage{epsfig}
\usepackage{amsmath,amssymb,amsfonts}
\usepackage{textcomp}
\usepackage{booktabs}
\usepackage{wrapfig}
\usepackage{multirow}
\usepackage{stfloats}
\usepackage{algorithm}
\usepackage{algorithmic}
\usepackage{subfig}

\usepackage{natbib}

\usepackage[hidelinks]{hyperref}

\title{FE-GUT: Factor Graph Optimization hybrid with Extended
Kalman Filter for tightly coupled GNSS/UWB Integration}

\author{
    Qijia~Zhao, \textit{Department of Precision Instrument, Tsinghua~University}
    \vspace{1mm} \\%
    Shaolin~Lü, \textit{State Key Laboratory of Precision Space-time Information Sensing Technology, Tsinghua~University}%
    \vspace{1mm} \\%
    Jianan~Lou, \textit{Department of Precision Instrument, Tsinghua~University}%
    \vspace{1mm} \\%
    Rong~Zhang*, \textit{State Key Laboratory of Precision Space-time Information Sensing Technology, Tsinghua~University}%
    }

\begin{document}

\maketitle

\section*{biography}


\biography{Qijia Zhao}{received the B.Eng. degree from Tsinghua University, Beijing, China, in 2023. He is pursuing his Ph.D. degree in the Department of Precision Instrument in Tsinghua University, Beijing, China. His researching interests focus on multi-sensor integrated navigation algorithms.}

\biography{Shaolin Lü}{received a PHD degree in navigation from Beijing Institute of Technology, Beijing, China, in 2009. From 2009 to
 2011, he was with Tsinghua University as a postdoctor. From 2011 to 2018, he was a senior engineer at the NORINCO group. From 2019 to 2024, he was the founder of Graph Optimization Inc. From September 2024, he is a research scientist at State Key Laboratory of Precision Space-time Information Sensing Technology, Tsinghua University. }

\biography{Jianan Lou}{received the B.Eng. degree in Navigation Engineering and the B.Ec. degree in Economics from Wuhan University, Wuhan, China, in 2022. He is currently pursuing his Ph.D. degree with the Department of Precision Instrument, Tsinghua University, Beijing, China. His research focuses on tightly coupled integration of GNSS/INS/Vision for precise positioning and mapping.}

\biography{Rong Zhang}{received the B.S. degree in mechanical design and manufacturing from Tsinghua University, Beijing, China,
 in 1992. He received the M.S. degree in precision instruments from Tsinghua University, and joined the faculty of Tsinghua
 University in 1994. In 2007, he received the PHD degree in precision instruments from Tsinghua University, and became
 a Professor in the Department of Precision Instrument, Tsinghua University. He is now the chief of State Key Laboratory of Precision Space-time Information Sensing Technology, Tsinghua University. His professional interests include the precision
 motion control system, MEMS inertial sensors, and integrated navigation systems.}

\section*{Abstract}

Precise positioning and navigation information has been increasingly important with the development of the consumer electronics market.
 Due to some deficits of Global Navigation Satellite System (GNSS), such as susceptible to interferences, integrating of GNSS with additional alternative sensors is
  a promising approach to overcome the performance limitations of GNSS-based localization systems.
   Ultra-Wideband (UWB) can be used to enhance GNSS in constructing an integrated localization system.
    However, most low-cost UWB devices lack a hardware-level time synchronization feature, which necessitates the estimation
     and compensation of the time-offset in the tightly coupled GNSS/UWB integration.
     Given the flexibility of probabilistic graphical models,
 the time-offset can be modeled as an invariant constant in the discretization of the continuous model.
 This work proposes a novel architecture in which Factor Graph Optimization (FGO) is hybrid with Extend Kalman Filter (EKF) for tightly coupled GNSS/UWB integration with online Temporal calibration (FE-GUT). FGO is utilized to precisely estimate the time-offset, while EKF provides initailization for the new factors and performs time-offset compensation. Simulation-based experiments validate the integrated localization performance of FE-GUT. In a four-wheeled robot scenario, the results demonstrate that, compared to EKF, FE-GUT can improve horizontal and vertical localization accuracy by 58.59\% and 34.80\%,
 respectively, while the time-offset estimation accuracy is improved by 76.80\%. All the source codes and
 datasets can be gotten via https://github.com/zhaoqj23/FE-GUT/.

\section{Introduction} \label{intro}

The continuous development of the consumer electronics market has stimulated an increasing demand for
 low-cost and precise positioning services \citep{taoMultiSensorFusionPositioning2022,caiLongRangeUWBPositioningBased2023,yangUAVWaypointOpportunistic2022}. 
 Naturally, GNSS receivers has become essential in modern society, providing high-precision, 
 24-hour, three-dimensional positioning information \citep{kaplanUnderstandingGPSGNSS2017}. Due to some deficits of GNSS, such as susceptible to interferences,
 the integration of GNSS and additional alternative sensors is a promising approach to overcome the performance limitations of GNSS-based localization systems.
 Ultra-Wideband (UWB), a carrier-less communication technology with nanosecond-level time resolution, offers excellent characteristics including high transmission rates,
  strong multipath capability, and interference resistance \citep{history,huangMultiGNSSPrecisePoint2022}. 
  Low-cost UWB devices can achieve centimeter-level ranging accuracy in harsh environments which sparks many interests in GNSS/UWB integration \citep{ultrawideband,InandOut}.
  \par

Temporal synchronization is crucial in multi-sensor fusion positioning architectures \citep{TimeSyn}.
 Ideally, a common triggering signal ensures all units work on the same clock at the hardware level. For instance, the Pulse-Per-Second (PPS) signal
  from the GNSS receiver can serve as the common reference clock for the multi-sensor network due to its precise and stable 
  timing characteristics \citep{ginstime,liGPSslavedTimeSynchronization2006}. Unfortunately, most pre-packaged UWB devices lack the synchronization feature.
   Consequently, software-based methods are often adopted to add timestamps into asynchronous measurements
   from different sensors \citep{Softtime}. However, clock shift, jitter, and the limited computational speed of navigation processors may
    introduce an unknown time-offset between the UWB timestamps and the true sampling instant. 
    Additionally, asynchronous sampling cannot be accurately approximated by interpolation in the tightly coupled integration, leading to temporal misalignment. 
    Therefore, online temporal calibration is highly required for tightly coupled GNSS/UWB integration. \par

Currently, EKF is regarded as the de facto benchmark algorithm for most state estimation problems \citep{KF}, which is widely 
employed in the positioning architectures \citep{IndoorPos,FastandRobust} and online temporal calibration in multi-sensor networks \citep{OnlineTemVio,TCUWBINS}.  
As an algorithm proposed in the 1960s, EKF cannot handle large-scale data per update. 
Traditionally, when using EKF, there was an implicit assumption that all the system states were discretized at the same frequency. 
However, this assumption may not be obeyed when FGO, 
which is capable of handling larger-scale data, is used \citep{factorgraph}. In many scenarios, different system states can
 be discretized at different frequencies from the continuous-time model \citep{ge_resolution_2008}.\par

Graphical State Space Model (GSSM) is a relatively novel method by which slowly changing or constant variables are discretized at a low frequency \citep{luGraphicalStateSpace2021}. 
This modeling approach transforms the dynamic Bayesian network into a multi-connected graph, enabling the inclusion of additional information and improved state 
estimation accuracy in certain cases when this approach fits the nature of system states very well. GSSM has successfully inspired a bunch of applications, 
such as Real-Time Kinematic (RTK) positioning \citep{Ge2022GlobalRP, yanRealtimeKinematicPositioning2023} and initial alignment \citep{ZHouInitialAlignment2022}.
\par

In the context of time-offset calibration for the tightly coupled GNSS/UWB problem, 
the time-offset between GNSS and UWB measurements remains nearly constant over a certain period since the computer clocks drift slowly.
 Therefore, the unknown, invariant time offset can be modeled as a constant in the sliding time window and the discrete state model will become a GSSM. 
 When the tightly coupled GNSS/UWB integration is modeled by GSSM and solved with FGO, 
 the constant time-offset estimation is significantly more precise and robust than that obtained by EKF. 
 However, it is frustrating that the accuracy of position and velocity estimation will be either equal to or sometimes even worse than that obtained by EKF. 
 These influences of different discretization methods are intriguing, perplexing and utterly important. We will elucidate the theorical reasons for these phenomena in future works.
  Currently, it is apparent that an architecture with a hybrid of FGO and EKF can effectively harness the advantages of both methods.\par

In this work, we propose a novel architecture in which FGO is hybrid with EKF for tightly coupled GNSS/UWB integration with online Temporal calibration (FE-GUT).
Simulation-based experiments compare the positioning and time-offset estimation performance of FE-GUT with the 
traditional EKF-based approach \citep{guoEnhancedEKFBasedTime2023}. The remainder of this paper is organized as follows. 
Section II introduces the traditional discrete-time state space model (TDTSSM) for  tightly coupled GNSS/UWB integration. Section III proposes GSSM for tightly coupled GNSS/UWB integration and details the FE-GUT procedure which bridges FGO and EKF. In Section IV, FE-GUT are compared with EKF in the positioning and time-offset estimation performance for a four-wheeled scenario. Conclusion and future works are presented in Section V.

\section{TRADITIONAL DISCRETE-TIME STATE SPACE MODEL FOR TIGHTLY COUPLED GNSS/UWB  INTEGRATION} \label{TDTSSM}
In the field of navigation, Kalman's methodology continues to hold a dominant position \citep{PaulKalman,DanKalman}. Firstly, build a continues-time state space model which comprises of system model and measurement model. If neccessary, linearize the models. Secondly, discretize the state space model in which all the state variables evolve at the same frequencey. Finally, EKF or other form of nonlinear filter will be utilized to solve the discrete-time state space model.
\par 
\subsection{Continues-Time State Space Model for Tightly Coupled GNSS/UWB  Integration}
The state vector for tightly coupled GNSS/UWB integration used in this paper consists of postion vector, velocity vector, acceleration vector, clock-shift, clock-drift and time-offset. 
It can be described as follows
\begin{equation} 
{X}= \begin{bmatrix}   {r}^T &   {v}^T &   {a}^T & \delta t & \delta f & t_{d} \end{bmatrix}^{T}
\end{equation}
The system model is a constant acceleration model and it goes as follows \citep{CA}
\begin{gather}
  \dot{X}=FX+q=\begin{bmatrix}
  0 & I_{3\times 3} & 0& 0&0&0\\
  0 & 0 & I_{3\times 3}& 0&0&0\\
  0 & 0 & 0& 0&0&0\\
  0 & 0 & 0& 0&1&0\\
  0 & 0 & 0& 0&0&0\\
  0 & 0 & 0& 0&0&0
  \end{bmatrix}X+q
  \end{gather}
Where $q \sim N(0,Q)$ is the process noise and $  {I}_{3\times 3}$ is the identity matrix with the third-order.
\par
In the state vector of the system, the time-offset $t_d$ between GNSS and UWB measurements comprises of two components: \\
(1) The misalignment between UWB timestamps and the actual sampling instants.\\
(2) The asynchronous sampling instants of both sensors. \\
The strategy for time-offset calibration involves modifying the UWB measurement model
 to compensate for the spatial displacement caused by the time-offset and its impact on UWB measurements. 
 The raw pseudorange and Doppler-shift measurements from GNSS receiver and the two-way range measurements from 
 UWB tag are directly utilized in the tightly coulped architecture. For simplicity, the translation extrinsics 
 between the GNSS receiver and the UWB tag are not considered in this part. The coordinates of all vectors are represented in the Earth-centered Earth-fixed (ECEF) coordinates.\par
The pseudorange measurement model can be expressed as follows \citep{kaplanUnderstandingGPSGNSS2017}
\begin{equation} \label{pse} \begin{aligned}  &P ^{\left ({i}\right)}_{G}= \underbrace {\sqrt {{\left ({x^{\left ({i}\right)}_{G}-r_{x}}\right)}^{2}+{\left ({y^{\left ({i}\right)}_{G}-r_{y}}\right)}^{2}+{\left ({z^{\left ({i}\right)}_{G}-r_{z}}\right)}^{2}}}_{\rho_{G}^{\left ({i}\right)}} + \delta t + \epsilon _{G}^{\left ({i}\right)}\end{aligned} \end{equation}
where\par
$P ^{\left ({i}\right)}_{G}$ is the pseudorange measurement;\par
$r = (r_{x},r_{y},r_{z})^T$ is the position vector of the GNSS receiver;\par
$(x^{\left ({i}\right)}_{G},y^{\left ({i}\right)}_{G},z^{\left ({i}\right)}_{G})^T$ is the position vector of the \emph{i}th satellite;\par
${\rho_{G}^{\left ({i}\right)}}$ is the true range between the GNSS receiver and the \emph{i}th satellite;\par
$\delta t$ is the ranging error resulting from the clock bias, which is also estimated and compensated in the integration architecture;\par
$\epsilon _{G}^{\left ({i}\right)}$ is the random noise in the pseudorange measurement.\par
The Doppler-shift measurement model is presented as follows \citep{kaplanUnderstandingGPSGNSS2017}
\begin{equation}  {D^{\left ({i}\right)}_{G}}={\dot {P }^{\left ({i}\right)}_{G}- {v}^{\left ({i}\right)}_{G}\cdot   {I}^{\left ({i}\right)}}
  = -   {I}^{\left ({i}\right)} \cdot   {v} + \delta f + \dot {\epsilon }_{G}^{\left ({i}\right)}\end{equation}

where\par
$D_{G}^{(i)}$ is the Doppler-shift measurement;\par
$v=(v_{x},v_{y},v_{z})^T$ is the velocity vector of the GNSS receiver;\par
$v^{\left ({i}\right)}_{G}$ is the velocity vector of the \emph{i}th satellite;\par
$\emph{I}^{(i)}$ is the unit LOS (line-of-sight) vector from the receiver to the \emph{i}th satellite;\par
$\delta f$ is the measuring error resulting from the clock drift;\par
$\dot{\epsilon}_{G}^{(i)}$ is the random noise in the Doppler-shift measurement.\par
On account of the time-offset $t_{d}$, the measurements of both sensors occur sequentially in the motion trajectory of the carrier with the GNSS receiver and the UWB tag. The constant acceleration model \citep{CA} is also utilized to compensate the UWB ranging error introduced by $t_{d}$. The modified UWB measurement model can be written as \citep{guoEnhancedEKFBasedTime2023}
\begin{equation}\begin{aligned} \label{uwb} P ^{\left ({j}\right)}_{U}=&\sqrt {\left ({x^{\left ({j}\right)}_{U}-\left ({r_{x}-v_{x}t_{d}-0.5a_{x}t^{2}_{d}}\right)}\right)^{2} +\,\left ({y^{\left ({j}\right)}_{U}-\left ({r_{y}-v_{y}t_{d}-0.5a_{y}t^{2}_{d}}\right)}\right)^{2}} \\&\overline {+\,\left ({z^{\left ({j}\right)}_{U}-\left ({r_{z}-v_{z}t_{d}-0.5a_{z}t^{2}_{d}}\right)}\right)^{2}} + \epsilon _{U}^{\left ({j}\right)}\end{aligned}\end{equation}
Where\par
$P_{U}^{(j)}$ is the measured two-way range from the receiver to the \emph{j}th UWB anchor;\par
$r = (r_{x},r_{y},r_{z})^T$ is the position vector of the UWB tag;\par
$v = (v_{x},v_{y},v_{z})^T$ is the velocity vector of the UWB tag;\par
$a = (a_{x},a_{y},a_{z})^T$ is the acceleration vector of the UWB tag;\par
$(x_{U}^{(j)},y_{U}^{(j)},z_{U}^{(j)})^T$ is the postion vector of the \emph{j}th UWB anchor;\par
$\epsilon_{U}^{(j)}$ is the random noise in the UWB measurement.\par
On the basis of the sensor models, the linearized measurement model goes as follows
\begin{equation}
Y = HX + \epsilon
\end{equation}
The measurement vector $Y$ of GNSS/UWB is written as
\begin{align} \label{y} \setlength{\arraycolsep}{0.5pt}   {Y}= \begin{bmatrix} P ^{\left ({1}\right)}_{G}&\cdots &P ^{\left ({n}\right)}_{G} &D^{\left ({1}\right)}_{G}&\cdots &D^{\left ({n}\right)}_{G} &P ^{\left ({1}\right)}_{U}&\cdots&P ^{\left ({m}\right)}_{U} \end{bmatrix}^{T} \end{align}\par
Where \emph{n} is the number of satellites within the field of vision and \emph{m} is the number of UWB anchors. $\epsilon \sim N(0, R)$ is the random noise vector.
The linearized measurement matrix $H$ can be derived from the first order Taylor expansion of the measurement models \citep{guoEnhancedEKFBasedTime2023}
\begin{align}\label{cjacobian} \setlength{\arraycolsep}{0.5pt}   {H}= \begin{bmatrix}   {H}^{G}_{n\times 3}&\quad   {0}_{n\times 3}&\quad   {0}_{n\times 3}&\quad   {1}_{n\times 1}&\quad   {0}_{n\times 1} &\quad   {0}_{n\times 1}\\   {0}_{n\times 3}&\quad   {H}^{G}_{n\times 3}&\quad   {0}_{n\times 3}&\quad   {0}_{n\times 1}&\quad   {1}_{n\times 1} &\quad   {0}_{n\times 1}\\   {H}^{UP}_{m\times 3}&\quad   {H}^{UV}_{m\times 3}&\quad   {H}^{UA}_{m\times 3}&\quad   {0}_{m\times 1}&\quad {0}_{m\times 1}&\quad  {H}^{UT}_{m\times 1}\\ \end{bmatrix}\end{align}
In (\ref{cjacobian}), the rows of ${H}^{G}_{n\times 3}$ are unit LOS vectors from the satellites to the GNSS receiver. It can be expressed as \citep{guoEnhancedEKFBasedTime2023}
\begin{align}  {H}^{G}_{n\times 3}=&
\begin{bmatrix} \dfrac {-\left ({x^{\left ({1}\right)}_{G}-r_{x}}\right)}{{\rho}_{G}^{\left ({1}\right)}}&\dfrac {-\left ({y^{\left ({1}\right)}_{G}-r_{y}}\right)}{{\rho}_{G}^{\left ({1}\right)}}&\dfrac {-\left ({z^{\left ({1}\right)}_{G}-r_{z}}\right)}{{\rho}_{G}^{\left ({1}\right)}}\\ \dfrac {-\left ({x^{\left ({2}\right)}_{G}-r_{x}}\right)}{{\rho}_{G}^{\left ({2}\right)}}&\dfrac {-\left ({y^{\left ({2}\right)}_{G}-r_{y}}\right)}{{\rho}_{G}^{\left ({2}\right)}}&\dfrac {-\left ({z^{\left ({2}\right)}_{G}-r_{z}}\right)}{{\rho}_{G}^{\left ({2}\right)}}\\ \vdots &\vdots &\vdots \\ \dfrac {-\left ({x^{\left ({n}\right)}_{G}-r_{x}}\right)}{{\rho}_{G}^{\left ({n}\right)}}&\dfrac {-\left ({y^{\left ({n}\right)}_{G}-r_{y}}\right)}{{\rho}_{G}^{\left ({n}\right)}}&\dfrac {-\left ({z^{\left ({n}\right)}_{G}-r_{z}}\right)}{{\rho}_{G}^{\left ({n}\right)}}\\ \end{bmatrix}
\end{align}\par

Similarly, ${H}^{UP}_{m\times 3}$, ${H}^{UV}_{m\times 3}$, ${H}^{UA}_{m\times 3}$ are the Jacobian matrices between the UWB ranges and the position, velocity and acceleration vectors of the UWB tag. ${H}^{UP}_{m\times 3}$ has a similar structure to ${H}^{G}_{n\times 3}$, whose rows are LOS vectors from the UWB anchors to the tag. ${H}^{UV}_{m\times 3}$, ${H}^{UA}_{m\times 3}$ and ${H}^{UT}_{m\times 1}$ are derived from the modified UWB measurement model (\ref{uwb})
\begin{align}  {H}^{UP}_{m\times 3}=&
\begin{bmatrix} \dfrac {-\left ({x^{\left ({1}\right)}_{U}-r_{x}}\right)}{{\rho}_{U}^{\left ({1}\right)}}&\dfrac {-\left ({y^{\left ({1}\right)}_{U}-r_{y}}\right)}{{\rho}_{U}^{\left ({1}\right)}}&\dfrac {-\left ({z^{\left ({1}\right)}_{U}-r_{z}}\right)}{{\rho}_{U}^{\left ({1}\right)}}\\ \dfrac {-\left ({x^{\left ({2}\right)}_{U}-r_{x}}\right)}{{\rho}_{U}^{\left ({2}\right)}}&\dfrac {-\left ({y^{\left ({2}\right)}_{U}-r_{y}}\right)}{{\rho}_{U}^{\left ({2}\right)}}&\dfrac {-\left ({z^{\left ({2}\right)}_{U}-r_{z}}\right)}{{\rho}_{U}^{\left ({2}\right)}}\\ \vdots &\vdots &\vdots \\ \dfrac {-\left ({x^{\left ({m}\right)}_{U}-r_{x}}\right)}{{\rho}_{U}^{\left ({m}\right)}}&\dfrac {-\left ({y^{\left ({m}\right)}_{U}-r_{y}}\right)}{{\rho}_{U}^{\left ({m}\right)}}&\dfrac {-\left ({z^{\left ({m}\right)}_{U}-r_{z}}\right)}{{\rho}_{U}^{\left ({m}\right)}}\\ \end{bmatrix}
\end{align}\par

\begin{equation}
	\begin{split}
		{H}^{UV}_{m\times 3} &= -{H}^{UP}_{m\times 3}t_{d} \\
	    {H}^{UA}_{m\times 3} &= -0.5{H}^{UP}_{m\times 3}t_{d}^2 \\
	    {H}^{UT}_{m\times 1} &= -H^{UP}_{m\times 3}({v}+{a}t_{d})
	\end{split}
\end{equation}\par

\subsection{TDTSSM of Tightly Coupled GNSS/UWB  Integration}
Traditionally, at the \emph{k}th epoch, the system state vector for tightly coupled GNSS/UWB  integration can be discretized as
\begin{equation} \label{xk}   {X}_{k}= \begin{bmatrix}   {r}^T_{k} &   {v}^T_{k} &   {a}^T_{k} & \delta t_{k} & \delta f_{k} & t_{d,k} \end{bmatrix}^{T}\end{equation}\par
The system model is discretized as
\begin{equation}
X_{k} = F_{k}X_{k-1} + q_{k}
\end{equation}
The discretized state transition matrix $F_{k}$ goes as
\begin{align} \label{fk} \setlength{\arraycolsep}{0.5pt}   {F}_{k} \!=\! \begin{bmatrix}   {I}_{3\times 3} \!&\quad   {I}_{3\times 3}\Delta t\! &\quad 0.5   {I}_{3\times 3}\Delta t^{2}\!&\quad   {0}_{3\times 1}\!&\quad   {0}_{3\times 1}\!&\quad   {0}_{3\times 1}\\   {0}_{3\times 3}\! &\quad   {I}_{3\times 3} \!&\quad   {I}_{3\times 3}\Delta t\!&\quad   {0}_{3\times 1}\!&\quad   {0}_{3\times 1}\!&\quad   {0}_{3\times 1}\\   {0}_{3\times 3}\! &\quad   {0}_{3\times 3} \!&\quad   {I}_{3\times 3}\!&\quad   {0}_{3\times 1}\!&\quad   {0}_{3\times 1}\!&\quad   {0}_{3\times 1}\\   {0}_{1\times 3} \!&\quad   {0}_{1\times 3} \!&\quad   {0}_{1\times 3}\!&\quad 1\!&\quad \Delta t\!&\quad 0\\   {0}_{1\times 3}\! &\quad   {0}_{1\times 3} \!&\quad   {0}_{1\times 3}\!&\quad 0\!&\quad 1\!&\quad 0\\   {0}_{1\times 3}\! &\quad   {0}_{1\times 3} \!&\quad   {0}_{1\times 3}\!&\quad 0\!&\quad 0\!&\quad 1\\ \end{bmatrix}\!\!\quad \!\!\!\!\!\!\!\!\! \end{align}\par
where $\Delta t$ is the time between consecutive epochs. The linearized measurement model is discretized as
\begin{equation}
Y_{k} = H_{k}X_{k} + \epsilon_{k}
\end{equation}
The measurement vector of GNSS/UWB at \emph{k}th epoch is written as
\begin{align} \label{yk} \setlength{\arraycolsep}{0.5pt}   {Y}_{k}= \begin{bmatrix} P ^{\left ({1}\right)}_{G,k}&\cdots &P ^{\left ({n}\right)}_{G,k} &D^{\left ({1}\right)}_{G,k}&\cdots &D^{\left ({n}\right)}_{G,k} &P ^{\left ({1}\right)}_{U,k}&\cdots&P ^{\left ({m}\right)}_{U,k} \end{bmatrix}^{T} \end{align}\par
The discrete measurement matrix is expressed as follows \citep{guoEnhancedEKFBasedTime2023}
\begin{align}\label{jacobian} \setlength{\arraycolsep}{0.5pt}   {H}_{k}= \begin{bmatrix}   {H}^{G}_{n\times 3,k}&\quad   {0}_{n\times 3}&\quad   {0}_{n\times 3}&\quad   {1}_{n\times 1}&\quad   {0}_{n\times 1}&\quad   {0}_{n\times 1}\\   {0}_{n\times 3}&\quad   {H}^{G}_{n\times 3,k}&\quad   {0}_{n\times 3}&\quad   {0}_{n\times 1}&\quad   {1}_{n\times 1} &\quad   {0}_{n\times 1}\\   {H}^{UP}_{m\times 3,k}&\quad   {H}^{UV}_{m\times 3,k}&\quad   {H}^{UA}_{m\times 3,k}&\quad   {0}_{m\times 1}&\quad   {0}_{m\times 1}&\quad   {H}^{UT}_{m\times 1,k}\\ \end{bmatrix}\end{align}\par
In (\ref{jacobian})
\begin{align} \label{hg}  {H}^{G}_{n\times 3,k}=&
\begin{bmatrix} \dfrac {-\left ({x^{\left ({1}\right)}_{G,k}-r_{x,k}}\right)}{{\rho}_{G,k}^{\left ({1}\right)}}&\dfrac {-\left ({y^{\left ({1}\right)}_{G,k}-r_{y,k}}\right)}{{\rho}_{G,k}^{\left ({1}\right)}}&\dfrac {-\left ({z^{\left ({1}\right)}_{G,k}-r_{z,k}}\right)}{{\rho}_{G,k}^{\left ({1}\right)}}\\ \dfrac {-\left ({x^{\left ({2}\right)}_{G,k}-r_{x,k}}\right)}{{\rho}_{G,k}^{\left ({2}\right)}}&\dfrac {-\left ({y^{\left ({2}\right)}_{G,k}-r_{y,k}}\right)}{{\rho}_{G,k}^{\left ({2}\right)}}&\dfrac {-\left ({z^{\left ({2}\right)}_{G,k}-r_{z,k}}\right)}{{\rho}_{G,k}^{\left ({2}\right)}}\\ \vdots &\vdots &\vdots \\ \dfrac {-\left ({x^{\left ({n}\right)}_{G,k}-r_{x,k}}\right)}{{\rho}_{G,k}^{\left ({n}\right)}}&\dfrac {-\left ({y^{\left ({n}\right)}_{G,k}-r_{y,k}}\right)}{{\rho}_{G,k}^{\left ({n}\right)}}&\dfrac {-\left ({z^{\left ({n}\right)}_{G,k}-r_{z,k}}\right)}{{\rho}_{G,k}^{\left ({n}\right)}}\\ \end{bmatrix}
\end{align}\par

\begin{align}  {H}^{UP}_{m\times 3,k}=&
\begin{bmatrix} \dfrac {-\left ({x^{\left ({1}\right)}_{U,k}-r_{x,k}}\right)}{{\rho}_{U,k}^{\left ({1}\right)}}&\dfrac {-\left ({y^{\left ({1}\right)}_{U,k}-r_{y,k}}\right)}{{\rho}_{U,k}^{\left ({1}\right)}}&\dfrac {-\left ({z^{\left ({1}\right)}_{U,k}-r_{z,k}}\right)}{{\rho}_{U,k}^{\left ({1}\right)}}\\ \dfrac {-\left ({x^{\left ({2}\right)}_{U,k}-r_{x,k}}\right)}{{\rho}_{U,k}^{\left ({2}\right)}}&\dfrac {-\left ({y^{\left ({2}\right)}_{U,k}-r_{y,k}}\right)}{{\rho}_{U,k}^{\left ({2}\right)}}&\dfrac {-\left ({z^{\left ({2}\right)}_{U,k}-r_{z,k}}\right)}{{\rho}_{U,k}^{\left ({2}\right)}}\\ \vdots &\vdots &\vdots \\ \dfrac {-\left ({x^{\left ({m}\right)}_{U,k}-r_{x,k}}\right)}{{\rho}_{U,k}^{\left ({m}\right)}}&\dfrac {-\left ({y^{\left ({m}\right)}_{U,k}-r_{y,k}}\right)}{{\rho}_{U,k}^{\left ({m}\right)}}&\dfrac {-\left ({z^{\left ({m}\right)}_{U,k}-r_{z,k}}\right)}{{\rho}_{U,k}^{\left ({m}\right)}}\\ \end{bmatrix}
\end{align}\par

\begin{equation}
  \label{hu}
	\begin{split}
		{H}^{UV}_{m\times 3,k} &= -{H}^{UP}_{m\times 3,k}t_{d,k} \\
	    {H}^{UA}_{m\times 3,k} &= -0.5{H}^{UP}_{m\times 3,k}t_{d,k}^2 \\
	    {H}^{UT}_{m\times 1,k} &= -H^{UP}_{m\times 3,k}({v}_{k}+{a}_{k}t_{d,k})
	\end{split}
\end{equation}\par
EKF can be utilized to solve the above problem \citep{PaulKalman,DanKalman}.\par
Initialize the estimation
\begin{gather}
\hat{X}_{0}=E(X_0)=X_0^+ \\
P_0=E[(X_0-\hat{X}_{0})(X_0-\hat{X}_{0})^T]
\end{gather}
Update the prior mean and covariance
\begin{gather}
X_{k|k-1}=F_kX_{k-1}\\
P_{k|k-1}=F_kP_{k-1}F_k^T+Q_k
\end{gather}
Calculate Kalman gain
\begin{gather}
K_{k}=P_{k|k-1}H_{k}^T(H_{k}P_{k|k-1}H_{k}^T+R_k)^{-1}
\end{gather}
Update the posteriori mean and covariance
\begin{gather}
X_{k}=X_{k|k-1}+K_{k}(Y_{k}-H(X_{k|k-1}))\\
P_{k}=(I-K_{k}H_{k})P_{k|k-1}
\end{gather}
Where $H(X_{k|k-1})$ represents the nonlinear measuring models described in (\ref{pse}) $\sim$ (\ref{uwb}).

\section{A New Hybrid Architecture for Tightly Coupled GNSS/UWB Integration}\label{gssm}
In this section, GSSM is introduced briefly and implemented. Whereafter, the GSSM implementation 
for tightly coupled GNSS/UWB  integration is given. A
novel architecture, in which FGO is hybrid with EKF for tightly coupled
GNSS/UWB integration with online Temporal calibration (FE-GUT), is proposed. 
\subsection{Graphical State Space Model for Tightly Coupled GNSS/UWB  Integration}
For some certain scenarios, the system may have state variables that are invariant over time. 
The system model can be rewritten as \citep{luGraphicalStateSpace2021}
\begin{gather}
  \dot{X}=
  \begin{bmatrix}
  \dot{X_c}  \\
  \dot{X_b}
  \end{bmatrix}
=FX+q=
\begin{bmatrix}
F_c & F_{b} \\
0 & 0
\end{bmatrix}
\begin{bmatrix}
X_c \\
X_b
\end{bmatrix}
+
\begin{bmatrix}
q_c \\
0
\end{bmatrix} \\
Y=HX+\epsilon=
\begin{bmatrix}
 H_c & H_b
 \end{bmatrix}
 \begin{bmatrix}
  X_c  \\
  X_b
  \end{bmatrix}+\epsilon
\end{gather}
where  $X_{b} \in R^{n_b}$  is time-invariant  and $X_{c}\in R^{n_c}$ is time-varying. 
\par
 Different from the discretization method which is described by (12)$\sim$(20), in GSSM setting for a $p$-size window, the above equation will be discretized as follows
\begin{gather} \label{tran} {X}_{w}(k) = \begin{bmatrix}   X_{c}(k) \\   X_{b}(k)\\\end{bmatrix}
   = \begin{bmatrix}  F_{w,c}(k)&  F_{w,b}(k) \\
{0} &\ {I} \\ \end{bmatrix}
\begin{bmatrix}   X_{c}(k-1) \\   X_{b}(k-1) \end{bmatrix}
+ \begin{bmatrix}   q_{w,c}(k) \\   0 \end{bmatrix} \\
  Y_{w}(k) = \begin{bmatrix}   H_{w,c}(k) &   H_{w,b}(k)\\\end{bmatrix} \begin{bmatrix}   X_{c}(k) \\   X_{b}(k)\\\end{bmatrix} +   \epsilon_{w,k} \end{gather}
where  $F_{w,c}(k)$ is the discretization of $F_c$ and  $F_{w,b}(k)$ is the discretization of $F_b$.
In the above equations, the system state vector of GSSM at \emph{k}th epoch can be written as
\begin{equation}   {X}_{w}(k)= \begin{bmatrix} {X}^T_{c}(k) & X^{T}_b(k)  \end{bmatrix}^{T} =
    \begin{bmatrix}  {X}^{T}_{k-p+1,c} &   {X}^{T}_{k-p+2,c} &\cdots&    {X}^{T}_{k-1,c} &   {X}^{T}_{k,c} &   X^{T}_{b}(k)
\end{bmatrix}^{T} \end{equation}
where  $X_{b}(k) \in R^{n_b}$  is the discretization of $X_{b}$ and $X_{c}(k) \in R^{p\times n_c}$ is the discretization of $X_{c}$. 
The measurement vector in the Equation (31) contains all of the sensor measurement in the $p$-size window, 
which can be written as
\begin{align}   Y_{w}(k)= \begin{bmatrix}
  Y^{T}_{k-p+1} &   Y^{T}_{k-p+2} & \cdots &   Y^{T}_{k-1} &   Y^{T}_{k} \end{bmatrix}^{T} \end{align}
where $Y_{k}\in R^{l}$ is the measurement vector at $k$th epoch.
\par
The linearized measurement matrix is defined as
\begin{gather} 
  H_{w,c}(k) =\begin{bmatrix}
    H_{c}(k-p+1) & 0_{l\times n_c} &\cdots&\cdots & 0_{l\times n_c} \\
    0_{l\times n_c}  &  H_{c}(k-p+2)& \cdots&\cdots & 0_{l\times n_c} \\
    \vdots &\vdots &\ddots &\vdots  & \vdots  \\
    0_{l\times n_c} & 0_{l\times n_c} & \cdots& H_{c}(k-1)& 0_{l\times n_c}\\
    0_{l\times n_c} &0_{l\times n_c}   & \cdots& \cdots &H_{c}(k)\\
    \end{bmatrix}\\
  H_{w,b}(k) = \begin{bmatrix} H_{b}(k-p+1) \\
     H_{b}(k-p+2)\\
    \vdots   \\
     H_{b}(k-1)\\
   H_{b}(k)\\\end{bmatrix}
\end{gather}
In this paper, for tightly coupled  GNSS/UWB integration, $X_b(k)$ represents the constant time-offset $t_{d,k}$ and $X_{c}(k)$ 
is the time-varying part in (\ref{xk}) at $k$th epoch:
\begin{align}   {X}_{c}(k) = \begin{bmatrix}   {r}^T_{k} &   {v}^T_{k} &   {a}^T_{k} & \delta t_{k} & \delta f_{k} \end{bmatrix}^{T}  \end{align} 

According to (\ref{fk}),
\begin{gather} 
  F_{w,b}(k)=0
  \end{gather} 
$F_{w,c}(k)$ represents the recursive relationships among the dynamic variables $X_c(k)$ within the sliding window, 
which can be formulated as follows
\begin{gather} 
  F_{w,c}(k) =\begin{bmatrix}
    F_{c}(k-p+1) & 0_{11\times 11} &\cdots&\cdots & 0_{11\times 11} \\
    0_{11\times 11}  &  F_{c}(k-p+2)& \cdots&\cdots & 0_{11\times 11} \\
    \vdots &\vdots &\ddots &\vdots  & \vdots  \\
    0_{11\times 11} & 0_{11\times 11} & \cdots& F_{c}(k-1)& 0_{11\times 11}\\
    0_{11\times 11} &0_{11\times 11}   & \cdots& \cdots &F_{c}(k)\\
    \end{bmatrix}
\end{gather}
where $F_{c}(k)$ is a submatrix of (\ref{fk})
\begin{align} \setlength{\arraycolsep}{0.5pt}   {F}_{c}(k) \!=\! \begin{bmatrix}   {I}_{3\times 3} \!&\quad   {I}_{3\times 3}\Delta t\! &\quad 0.5   {I}_{3\times 3}\Delta t^{2}\!&\quad   {0}_{3\times 1}\!&\quad   {0}_{3\times 1}\\   {0}_{3\times 3}\! &\quad   {I}_{3\times 3} \!&\quad   {I}_{3\times 3}\Delta t\!&\quad   {0}_{3\times 1}\!&\quad   {0}_{3\times 1}\\   {0}_{3\times 3}\! &\quad   {0}_{3\times 3} \!&\quad   {I}_{3\times 3}\!&\quad   {0}_{3\times 1}\!&\quad   {0}_{3\times 1}\\   {0}_{1\times 3} \!&\quad   {0}_{1\times 3} \!&\quad   {0}_{1\times 3}\!&\quad 1\!&\quad \Delta t\\   {0}_{1\times 3}\! &\quad   {0}_{1\times 3} \!&\quad   {0}_{1\times 3}\!&\quad 0\!&\quad 1 \end{bmatrix}\!\!\quad \!\!\!\!\!\!\!\!\! \end{align}
The components in $Y_k$ is the same as (\ref{yk}), which is presented for legibility
\begin{align} \label{yk} \setlength{\arraycolsep}{0.5pt}   {Y}_{k}= \begin{bmatrix} P ^{\left ({1}\right)}_{G,k}&\cdots &P ^{\left ({n}\right)}_{G,k} &D^{\left ({1}\right)}_{G,k}&\cdots &D^{\left ({n}\right)}_{G,k} &P ^{\left ({1}\right)}_{U,k}&\cdots&P ^{\left ({m}\right)}_{U,k} \end{bmatrix}^{T} \end{align}
The linearized measurement matrix $H_c(k)$ and $H_b(k)$ for the \emph{k}th epoch are submatrices of $H_{k}$ in (\ref{jacobian})
\begin{gather}
\setlength{\arraycolsep}{0.5pt}   {H}_{c}(k)= \begin{bmatrix}   {H}^{G}_{n\times 3,k}&\quad   {0}_{n\times 3}&\quad   {0}_{n\times 3}&\quad   {1}_{n\times 1}&\quad   {0}_{n\times 1}\\   {0}_{n\times 3}&\quad   {H}^{G}_{n\times 3,k}&\quad   {0}_{n\times 3}&\quad   {0}_{n\times 1}&\quad   {1}_{n\times 1}\\   {H}^{UP}_{m\times 3,k}&\quad   {H}^{UV}_{m\times 3,k}&\quad   {H}^{UA}_{m\times 3,k}&\quad   {0}_{m\times 1}&\quad   {0}_{m\times 1}\\ \end{bmatrix} \\
\setlength{\arraycolsep}{0.5pt}   {H}_{b}(k)= \begin{bmatrix} 0_{2n \times 1} \\ H_{m \times 1,k}^{UT} \end{bmatrix}
\end{gather}
The terms mentioned in the measurement matrix can be calculated by the Equations (\ref{hg})$\sim$(\ref{hu}).

\subsection{The FE-GUT Architecture}



\begin{figure}[thpb]
      \centering
      \includegraphics[width=0.8\linewidth]{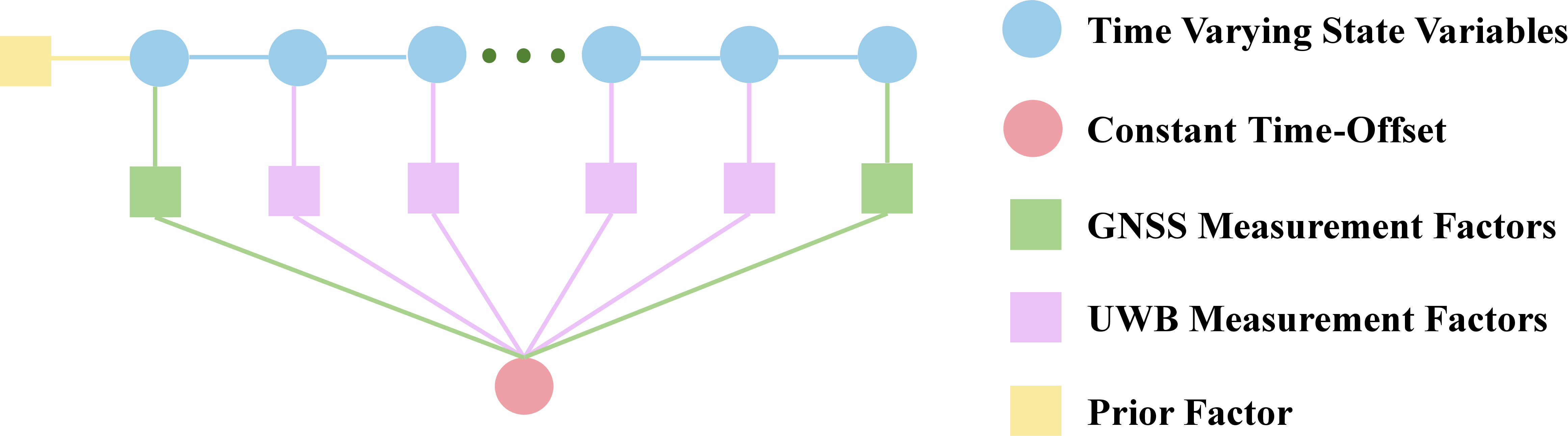}
      \caption{The factor graph of tightly coupled GNSS/UWB integration modeled by GSSM.}
      \label{fg}
\end{figure}

Ceres Solver \citep{Agarwal_Ceres_Solver_2022}, an open-source C++ library developed by Google, is used to solve this nonlinear optimization
 problem modeled by GSSM. The factor graph framework of GSSM for tightly coupled GNSS/UWB  integration is shown in Fig. \ref{fg},
  which consists of the time-varying state variables, the constant time-offset variable, the prior factor, state prediction factors,
   GNSS measurement factors and UWB measurement factors. The prior factor is an initial prior estimation of the system states. 
   Via marginalization, the prior factor contains historical constraint information when the sliding-window graph optimization is utilized. 
   The state prediction factors are binary factors modeled to establish the state transition constraint between two adjacent epochs. 
   The GNSS and UWB measurement factors can be constructed through the sensor measurements which models are described in Section \ref{TDTSSM}.

\begin{figure}[thpb]
      \centering
      \includegraphics[width=0.6\linewidth]{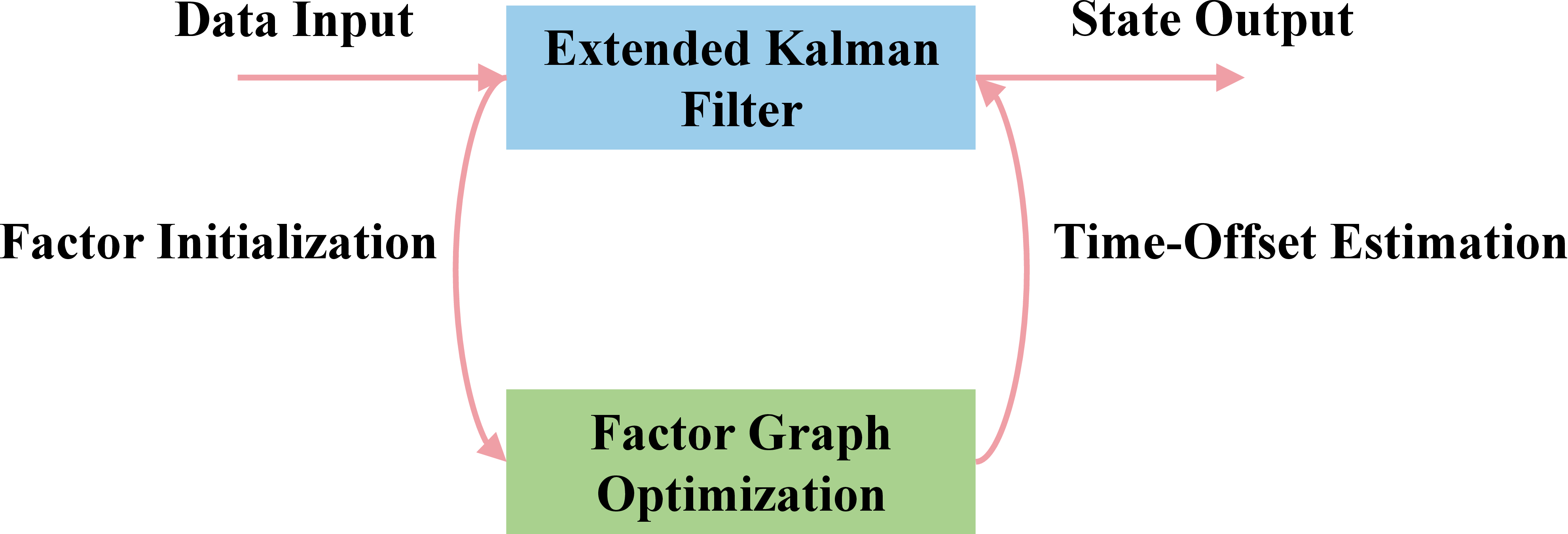}
      \caption{The architecture in which FGO is hybrid with EKF for tightly coupled GNSS/UWB integration.}
      \label{arc}
\end{figure}

As mentioned in Section \ref{intro}, GSSM can perform better in the time-offset estimation but have equal or even worse accuracy in results of positioning. 
To tackle this problem, the time-offset obtained from FGO will be fed back to a naive EKF, which lacks the function of temporal calibration, to compensate for the time-offset. 
Moreover, EKF can be used to initialize the new factors added in the sliding window. This initialization can provide an initial guess located near the global 
optimum, preventing the optimization result from falling into the local optimum, and thus improving the speed and accuracy of convergence. The block diagram of 
the algorithm can be depicted in Fig. \ref{arc}.
\par
The algorithm pseudocode is presented in Algorithm \ref{alo}, which C++ implementation can be found at GitHub. It is northworthy that the UWB devices have a relatively high sampling rate than GNSS. Consequently,
 the integration can degenerate into UWB localization by adjusting the dimensions of the measurement vector and the measurement matrix to include only UWB ranges at the \emph{k}th epoch.

\begin{algorithm}
\caption{A single update of FE-GUT for tightly coupled GNSS/UWB integration}\label{alo}
\begin{algorithmic}[1]
\REQUIRE Pseudorange $P^{(1 \sim n)}_{G,k}$; Doppler-shift $D^{(1 \sim n)}_{G,k}$; UWB range $P^{(1 \sim m)}_{U,k}$; position of Satellites $r^{(1 \sim n)}_{G,k}$; position of UWB anchors $r^{(1 \sim m)}_{U,k}$
\STATE State initialization;
\IF{Only have UWB data}
    \STATE EKF partial update($P^{(1 \sim n)}_{G,k}$, $r^{(1 \sim m)}_{U,k}$)
    \RETURN
\ELSE
    \STATE EKF general update($P^{(1 \sim n)}_{G,k}$, $D^{(1 \sim n)}_{G,k}$, $P^{(1 \sim m)}_{U,k}$, $r^{(1 \sim n)}_{G,k}$, $r^{(1 \sim m)}_{U,k}$)
\ENDIF
\STATE Factor Graph Optimization with states initialized by EKF
\STATE Marginalization
\STATE Sliding window
\STATE Time-offset feedback
\RETURN
\ENSURE Position $r_{k}$; velocity $v_{k}$; time-offset $t_{d,k}$
\end{algorithmic}
\end{algorithm}

\section{Results and Analysis}
In this section, simulation-based experiments are conducted to validate the integrated localization performance of FE-GUT.
 In Subsection \ref{simulationset}, the simulation setup and methodology are introduced.
  Then, in Subsection \ref{simulationresult}  the experiment results of FE-GUT versus EKF are given.
\subsection{Simulation Setup} \label{simulationset}
The simulation scenario is a four-wheeled robot. The average speed of the robot is $5m/s$ and the norminal $t_d$ is set to $40ms$. 
The four UWB anchors are symmetrically placed around the center of the trajectory, with a horizontal distance of $50m$ and a height of $5m$. 
An elevation mask of $15^{\circ}$ is set to avoid GNSS positioning performance degradation caused by low elevation satellites. 
The UWB ranges are construced by a time-of-arrival (TOA) ranging model. Gaussian white noise with standard deviations of $2m$, $0.1m/s$, and $0.1m$ are added to the measurements of GNSS pseudorange, Doppler-shift and UWB range, respectively.\par

The UWB dataset is pre-processed to be time-misaligned with the GNSS dataset based by the time-offset $t_d$. In detail, we generate a high temporal resolution motion information look-up table, where interpolation can be performed to obtain the position at any time and calculate the GNSS and UWB measurements. The UWB measurement always has a time lag of $t_d$ compared with the GNSS and the positioning error increases due to the integration of asynchronous data.
 When the receiver has high kinematics, the uncalibrated $t_d$ can contribute a large proportion of errors compared with other sources.\par

The simulation trajectory is the Bernoullian Lemniscate with the major axis parallel to the east-west direction and a horizontal extension of $200m$. The western apex is in a location of latitiude $39.904987^{\circ}$, longtitude $116.405289^{\circ}$ and height $60.0352m$, specified by the WGS84 coordinates. Additionally simulation with circular trajectory is also conducted. Due to space limitation, only the error statistical results are presented.\par

\subsection{State-Estimation Performance} \label{simulationresult}
\begin{figure*}[htbp]
  \centering
  \subfloat
  {\label{pos}\includegraphics[width=0.5\textwidth]{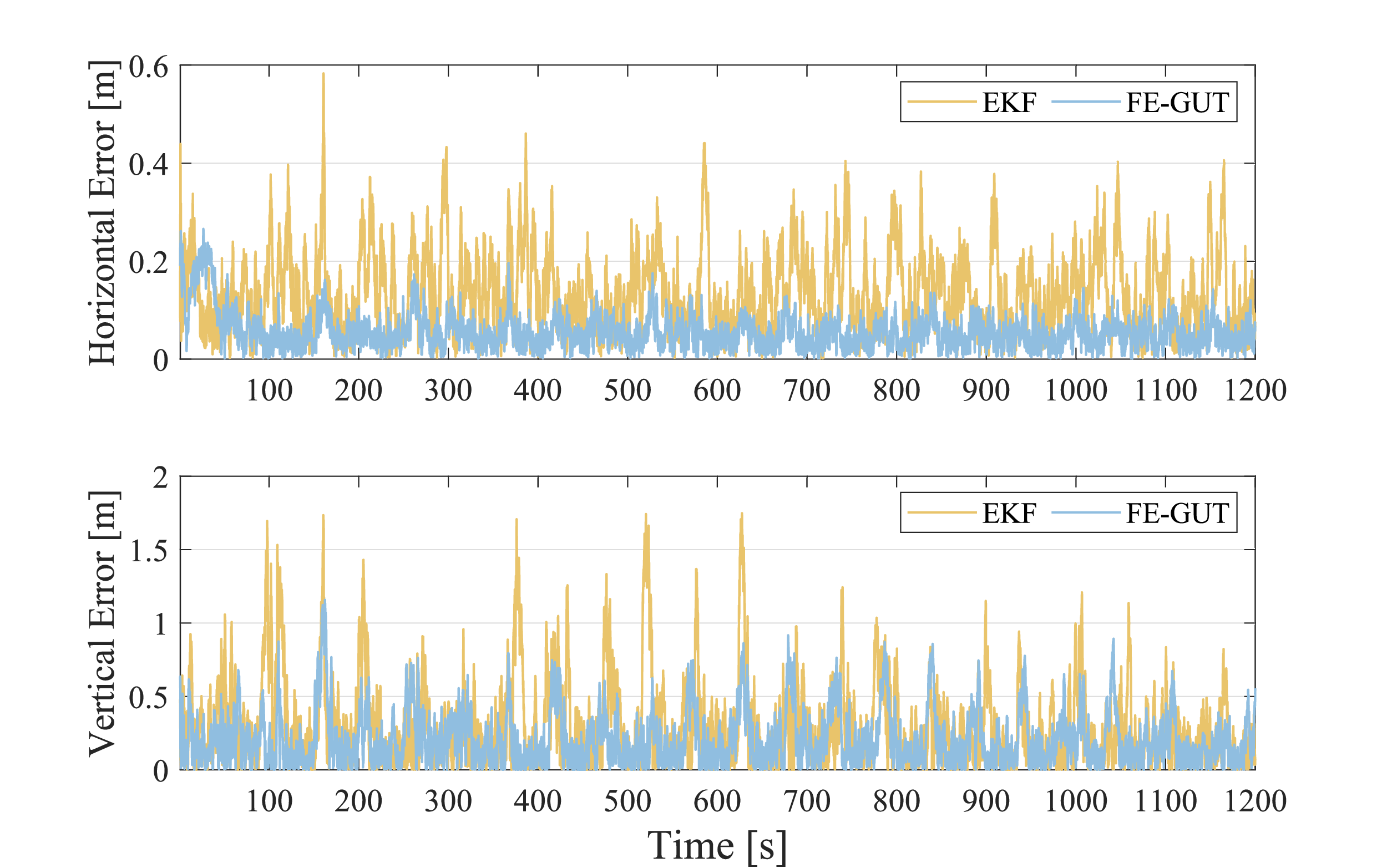}}
  \subfloat
  {\label{td}\includegraphics[width=0.5\textwidth]{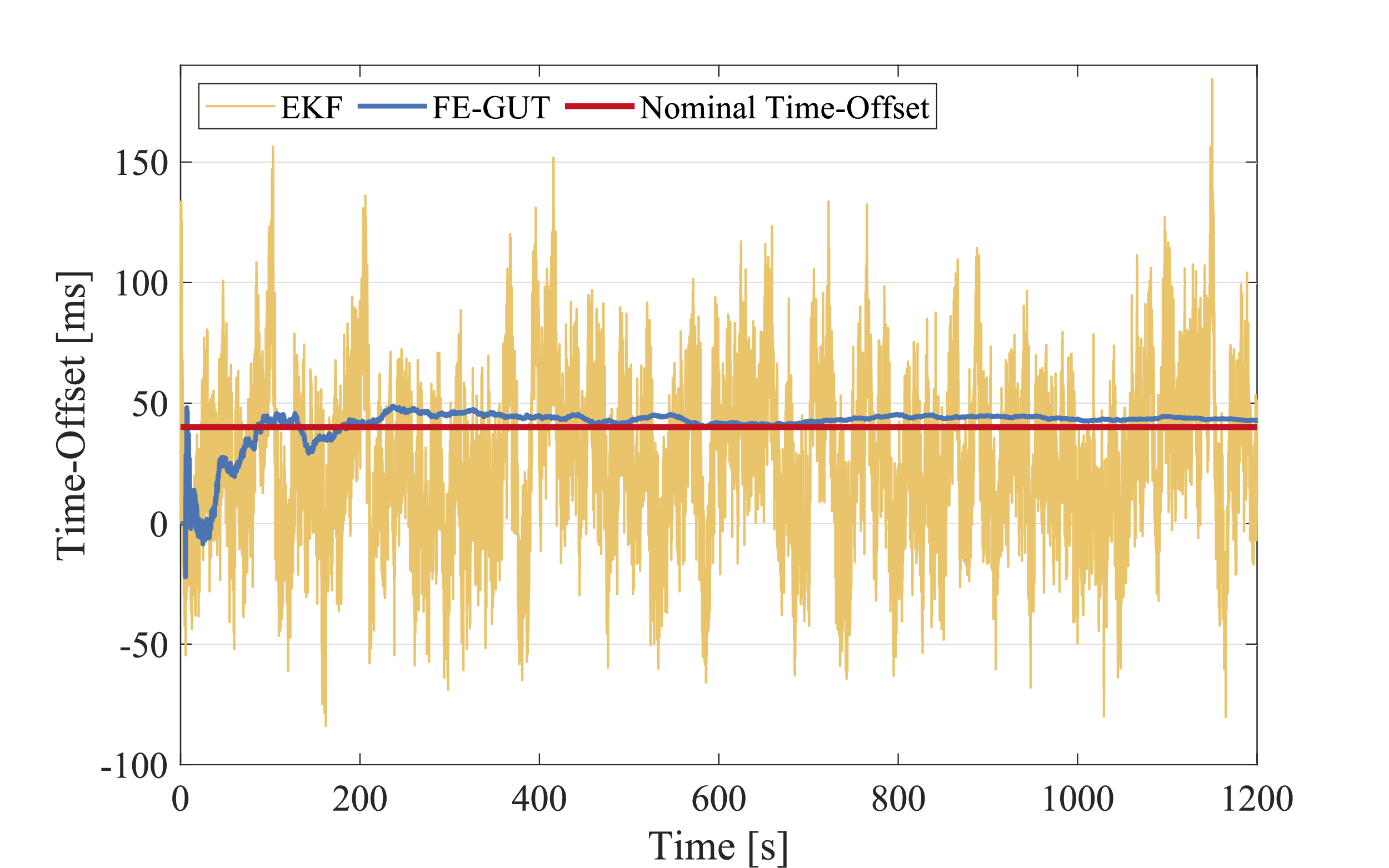}}
  \newline
  \subfloat
  {\label{traj}\includegraphics[width=0.8\textwidth]{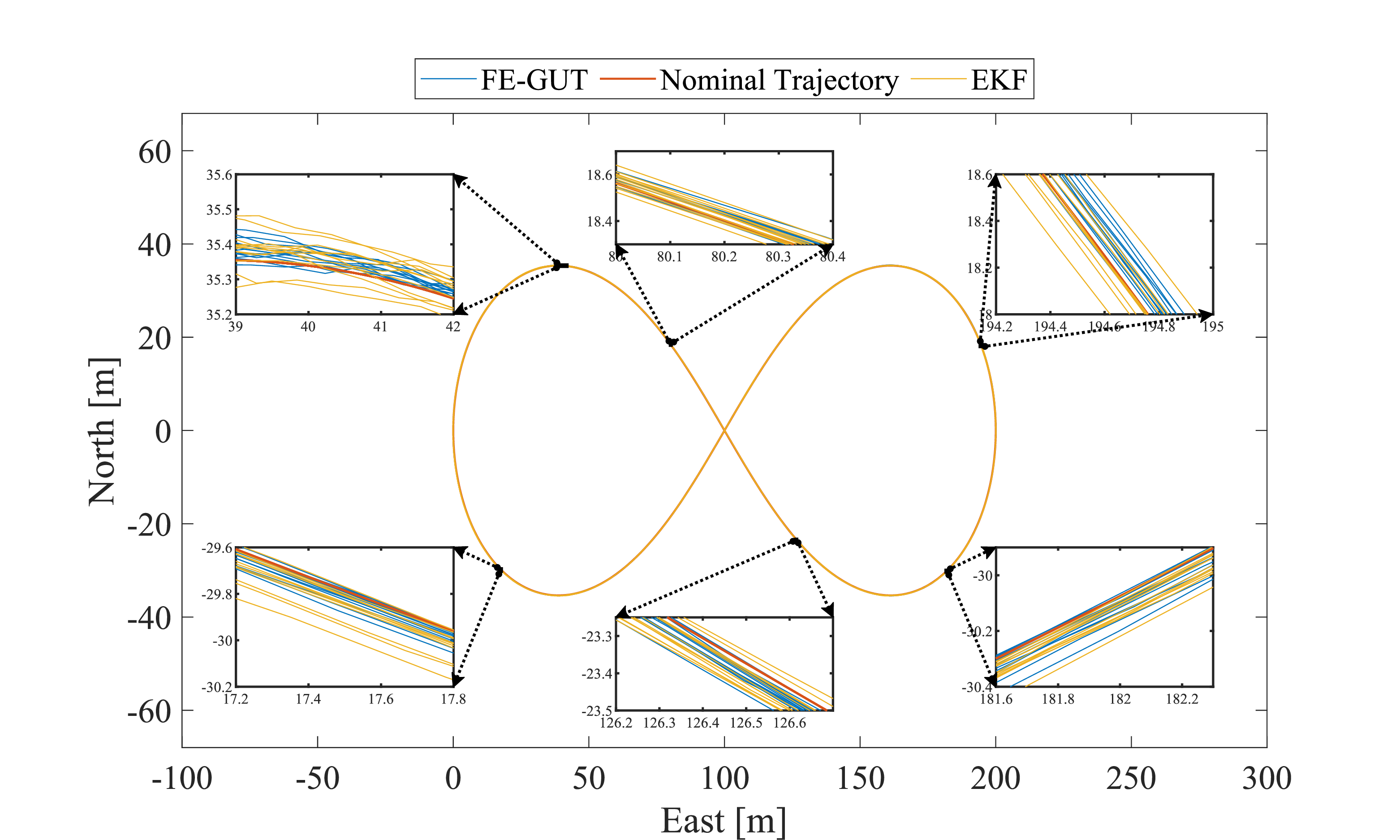}}
  \caption{(a) The comparison of the horizontal and vertical positioning localization error between EKF and FE-GUT. (b) The comparative evaluation of the estimated time offset between EKF and FE-GUT. (c) The nominal and estimated trajectories.}
  \label{state-estimation}
\end{figure*}

In Fig. \ref{state-estimation}, the positioning error and the time series of $t_d$ estimation obtained from EKF and FE-GUT are compared.
In Fig. \ref{pos}, the positioning error in both horizontal and vertical directions is compared for FE-GUT and EKF. In Fig. \ref{td},
the time-offset estimation results from FE-GUT are compared with those from EKF. In Fig. \ref{traj}, the nominal and estimated trajectories are illustrated. \par

It can be clearly observed that FE-GUT significantly improves the positioning accuracy. The RMSE of estimated positioning and time-offset results are presented in Table \ref{estimation}. For the Bernoullian Lemniscate trajectory, FE-GUT can improve the horizontal and vertical positioning accuracy 
by $58.59\%$ and $34.80\%$, respectively, compared to EKF, as indicated by the RMSE metric. This improvement is predominantly due to the enhancement in
time-offset estimation accuracy, which is $76.80\%$. In addition to statistics, the improvement of state estimation performance is also highlighted in Fig. \ref{td} and Fig. \ref{td}. Fig. \ref{traj} 
illustrates that the results obtained from FE-GUT are more smoother and closer to the nominal trajectory. 
\par

It is noteworthy that the main purpose of this work is to verify the capacity of GSSM in temporal calibration. The position of UWB anchors is not specifically designed for 3D positioning (the height of the anchors is all $5m$), which lead to relatively poor geometric distribution of UWB anchors. Therefore, the vertical positioning accuracy obtained from FE-GUT and EKF is worse than the results in horizontal direction. Addtionally, the simulation scenario involves a four-wheeled scene where the vertical velocity is near to zero, which means the vertical spatial misalignment caused by time-offset is also negligible. The improvement of time-offset estimation has tiny impact on the vertical positioning. Essentially, the main contribution of FE-GUT is to enhance the $t_d$ estimation performance, which may improve the positioning accuracy.

\begin{table*}[htbp]
    \centering
    \caption{State-estimation RMSE of FE-GUT and EKF}
    \label{estimation}
    \begin{tabular}{ccccccc}
        \toprule
        \multirow{2}*{Integration Architecture} & \multicolumn{3}{c}{Bernoullian Lemniscate} & \multicolumn{3}{c}{Circle} \\
        \cmidrule(lr){2-4} \cmidrule(lr){5-7}
        & Horizontal(m) & Vertical(m) & Time-offset(ms) & Horizontal(m) & Vertical(m) & Time-offset(ms) \\
        \midrule
        EKF   & 0.158  & 0.425  & 35.656  & 0.124  & 0.196  & 32.26 \\
        \midrule
        FE-GUT  & 0.065  & 0.277  & 8.271 & 0.056  & 0.190  & 10.168 \\
        \midrule
        Enhancement(\%) & 58.59 & 34.80 & 76.80 & 55.38 & 2.94 & 68.48\\
        \bottomrule
    \end{tabular}
\end{table*}

\section{Conclusion and future works}
In this work, the time-offset between GNSS and UWB are precisely estimated and compensated in the FE-GUT architecture. Low-cost UWB devices can be integrated with GNSS receivers for high-accuracy positioning with online temporal calibration. Addtionally, the effectiveness of GSSM is further demonstrated and its application range is expanded. Although GSSM solved by FGO can not outperform EKF for all variables in this context, it can still be used to augment EKF. Generally, GSSM enhances the estimation accuracy of constants in the system model, which is a feature worth leveraging. Besides, FGO and EKF are not in conflict with each other. The combination of these two algorithms can sometimes yield a brand-new and high-accuracy solution. In the future, real-world experiments will be conducted to validate the performance of FE-GUT. Regarding GSSM, more application scenarios will be explored and we will also attempt to elucidate the theorical reasons of changes in estimation accuracy resulting from GSSM.

\section*{acknowledgements}
We would like to acknowledge Mr.Yihan Guo for his kind help in this work.

\bibliographystyle{apalike}
\bibliography{GNSS-UWB.bib}

\end{document}